%% file: arxiv.tex
\documentclass[10pt,twocolumn,letterpaper]{article}

 \usepackage{iccv}              
\usepackage{multirow}

\definecolor{iccvblue}{rgb}{0.21,0.49,0.74}
\usepackage[pagebackref,breaklinks,colorlinks,allcolors=iccvblue]{hyperref}

\def\paperID{9763} 
\def\confName{ICCV}
\def\confYear{2025}

\title{DecoFuse: Decomposing and Fusing the ``What'', ``Where'', and ``How'' for Brain-Inspired fMRI-to-Video Decoding}

\author{Chong Li,
Jingyang Huo, Weikang Gong, Yanwei Fu, Xiangyang Xue, and Jianfeng Feng\\
Fudan University\\
{\tt\small lichong23@m.fudan.edu.cn}\\
}
\begin{document}
\maketitle

\begin{abstract}

	%
	%
	%
	%
	
	Decoding visual experiences from brain activity is a significant challenge. Existing fMRI-to-video methods often focus on semantic content while overlooking spatial and motion information. However, these aspects are all essential and are processed through distinct pathways in the brain. Motivated by this, we propose \textbf{DecoFuse}, a novel brain-inspired framework for decoding videos from fMRI signals. It first decomposes the video into three components\textemdash semantic, spatial, and motion\textemdash then decodes each component separately before fusing them to reconstruct the video. This approach not only simplifies the complex task of video decoding by decomposing it into manageable sub-tasks, but also establishes a clearer connection between learned representations and their biological counterpart, as supported by ablation studies. Further, our experiments show significant improvements over previous state-of-the-art methods, achieving 82.4\% accuracy for semantic classification, 70.6\% accuracy in spatial consistency, a 0.212 cosine similarity for motion prediction, and 21.9\% 50-way accuracy for video generation. Additionally, neural encoding analyses for semantic and spatial information align with the two-streams hypothesis, further validating the distinct roles of the ventral and dorsal pathways. Overall, DecoFuse provides a strong and biologically plausible framework for fMRI-to-video decoding. Project page: https://chongjg.github.io/DecoFuse/.

	
	%
	%
	
\end{abstract}

\begin{figure*}[t]
	\centering
	\includegraphics[width=0.8\linewidth]{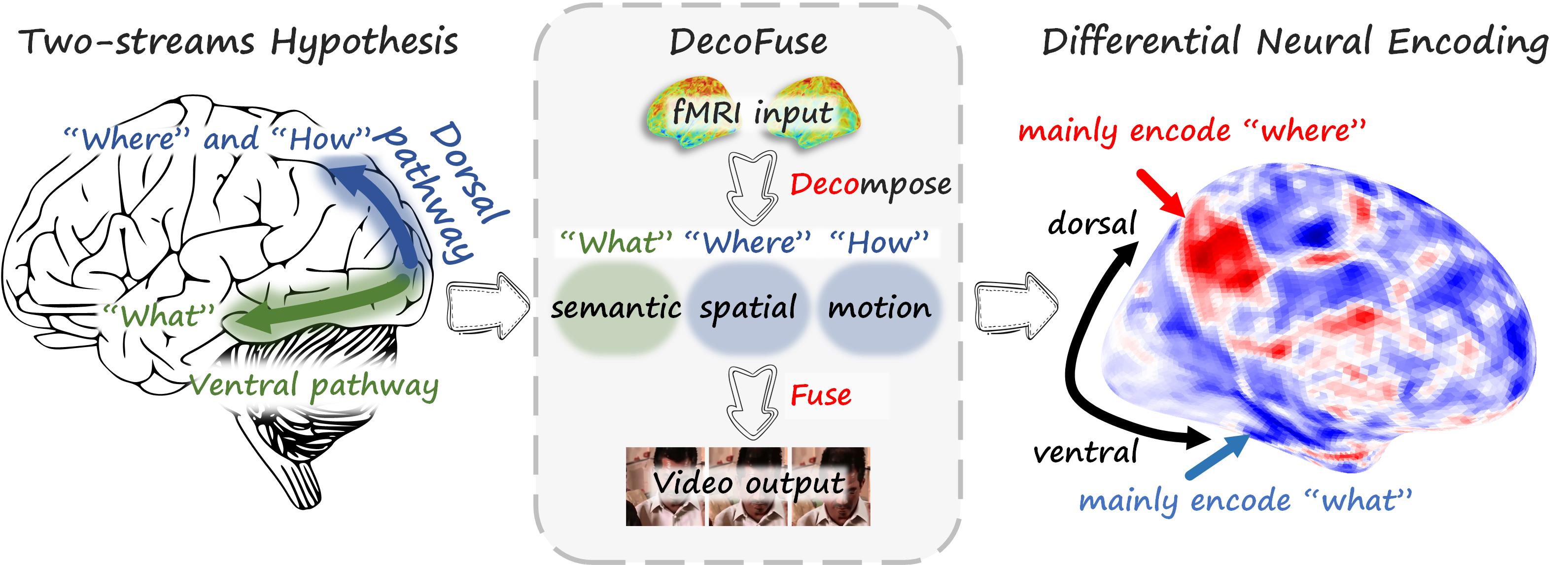}
	\vspace{-0.1in}
	\caption{\textbf{Diagram of DecoFuse framework.} Inspired by the brain’s two-streams hypothesis~\cite{GOODALE199220}, the \textbf{DecoFuse} pipeline decomposes video into three components: semantic (``what''), spatial (``where''), and motion (``how''). Neural features are extracted by an fMRI encoder and decomposed to semantic, spatial and motion embeddings. These components are then fused to generate video. Additionally, neural encoding analyzes the differential contribution of semantic and spatial embeddings in predicting signals from the brain's dorsal and ventral streams, confirming alignment with the two-streams hypothesis~\cite{GOODALE199220}.\label{fig:diagram}}
	\vspace{-0.15in}
\end{figure*}

\begin{figure}[t]
	\centering
	\includegraphics[width=0.9\linewidth]{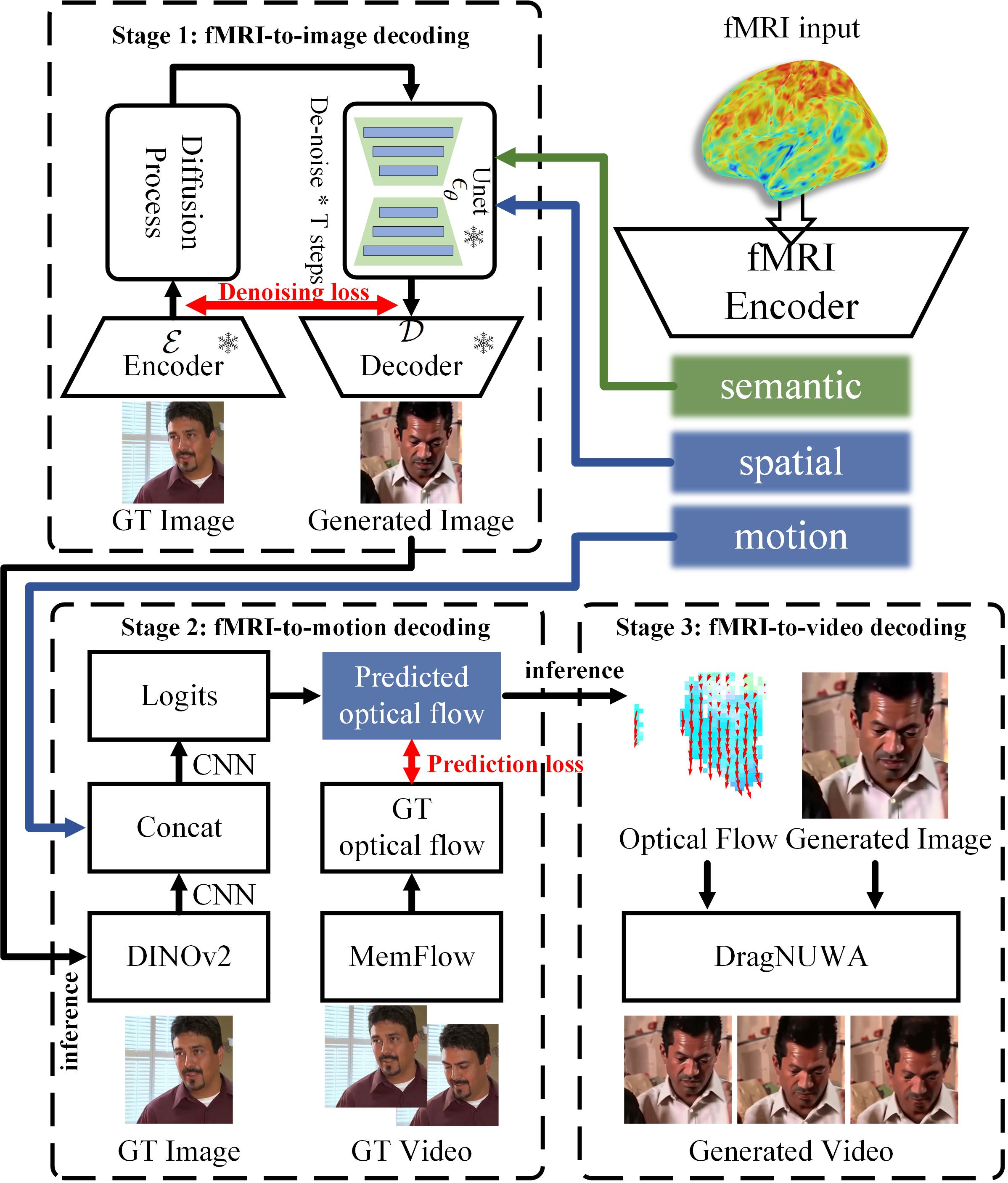}
	\vspace{-0.1in}
	\caption{\textbf{Details of DecoFuse framework.}
		Neural features are extracted by an fMRI encoder and decomposed to semantic, spatial and motion embeddings through three independent encoders. These components are then fused to generate video via three stages: \textbf{(1) fMRI-to-image decoding}, which uses Stable Diffusion and ControlNet to generate static images based on high-level semantic and low-level spatial embeddings; \textbf{(2) fMRI-to-motion decoding}, predicting optical flow using an image- and fMRI-based motion decoder to capture dynamic elements of the video; \textbf{(3) fMRI-to-video decoding}, where the decoded image and optical flow are combined to generate the final video using a motion-conditioned video diffusion model.  \label{fig:method}}
	\vspace{-0.15in}
\end{figure}

\section{Introduction}

Visual input is the brain’s primary source of information, making the accurate decoding of visual signals and understanding their encoding processes key challenges in neuroscience and AI. Functional magnetic resonance imaging (fMRI), a non-invasive method for recording whole-brain activity, has become increasingly popular for decoding applications~\cite{tong2012decoding}. Meanwhile, advances in techniques like Stable Diffusion (SD)\cite{SD} have driven major progress in fMRI-based decoding for images~\cite{qian2023joint,mind-vis,brain-diffuser,mind-reader,brain-clip,huo2025neuropictor}, videos~\cite{lienhancing,mind-video,gong2024neuroclips,BrainNetflix}, and 3D objects~\cite{mind-3d}. These breakthroughs have delivered remarkable results, bringing the idea of ``mind reading'' closer to reality.


However, decoding fMRI into video is still inherently  challenging!
Neuroscience research has shown that different brain regions process various aspects of visual information. The two-streams hypothesis~\cite{GOODALE199220, milner2006visual} suggests two main pathways for visual processing: the ``what'' pathway (ventral stream) for object recognition and the ``where/how'' pathway (dorsal stream) for tracking location and movement. These three components—semantic (what), spatial (where), and motion (how)—are fundamental to video perception. However, fMRI-to-video decoding has mainly focused on semantic information, while decoding spatial and motion aspects, which are crucial for visual experiences, remains a significant yet underexplored challenge~\cite{gong2024neuroclips}.

MinD-Video~\cite{mind-video} was the first to use Stable Diffusion for fMRI-to-video decoding, aligning fMRI features with text embeddings to reconstruct semantically accurate videos. Several studies have since followed this approach, focusing on semantic alignment~\cite{lienhancing, sun2024neurocine}. Yeung et al.~\cite{yeung2024neural} took a different approach by successfully decoding visual motion information.
Particularly,
recent works have also explored spatial decoding by predicting the variational autoencoder (VAE) latent of Stable Diffusion as an initial estimate for UNet’s noise input~\cite{lu2024animate, foscobrain, gong2024neuroclips}. Despite these efforts, evaluations mostly rely on semantic or pixel-level metrics like classification accuracy and SSIM. \textit{How well spatial and motion information can be independently decoded from fMRI remains an open question.}

To address these issues, we introduce DecoFuse, a novel brain-inspired framework that decomposes video into three key components—semantic, spatial, and motion information. They are separately decoded and then fused to reconstruct the video in \cref{fig:diagram}. Aligned with  two-streams hypothesis, the learned components are expected to reflect their biological counterparts in the brain as three stages: \\
\noindent \textbf{Stage 1}: A pretrained fMRI encoder extracts semantic, spatial, and motion embeddings. Semantic and spatial embeddings then condition an image generator, defining ``what'' the object is and ``where'' it is located, producing a static initial frame. \\
\noindent \textbf{Stage 2}: The motion decoder predicts optical flow using the neural motion embedding and the initial frame, simulating how the brain processes object movement. \\
\noindent  \textbf{Stage 3}: A motion-conditioned video generator animates the static frame using the predicted optical flow. 

DecoFuse offers two main advantages: (1) It simplifies fMRI-to-video decoding by breaking it into manageable sub-tasks, enhancing performance, and (2) its biologically inspired modular design supports ablation studies, allowing assessment of how well semantic, spatial, and motion information can be independently decoded from fMRI signals.


In our experiments, we evaluated decoding accuracy for each of the three components (semantic, spatial, and motion) and demonstrated superior performance compared to existing SOTA methods~\cite{mind-video,qian2023joint,lienhancing,yeung2024neural,huo2025neuropictor}. For semantic information, we conducted a classification task on the generated and ground truth (GT) images, achieving 20.8\% 50-way accuracy\textemdash an improvement of 20.9\% over MinD-Video~\cite{mind-video}. For spatial information, we applied foreground detection using DINOv2~\cite{oquab2023dinov2} and obtained a 70.6\% accuracy for foreground consistency between generated and GT images, surpassing the previous SOTA performance of 68.7\% in NeuroPictor~\cite{huo2025neuropictor}. Regarding motion information, we measured the cosine similarity between the predicted and GT optical flow, achieving a score of 0.212, significantly better than the 0.174 reported by \cite{yeung2024neural}. Moreover, we also assessed the quality of the generated videos, showing 50-way classification accuracy of 21.9\%, which outperforms current SOTA methods~\cite{mind-video,qian2023joint,lienhancing}. We also conducted ablation studies for each component, all of which showed a significant drop in their respective metrics, emphasizing the correspondence between our learned representations and their biological counterparts. Finally, leveraging our brain-inspired decomposition in DecoFuse, we conducted neural encoding of ``what'' and ``where'' embeddings, demonstrating alignment with the two-streams hypothesis~\cite{GOODALE199220}.

In summary, we have these contributions: 
\textbf{(1) \textit{Novel Brain Decoding Framework}}:
This paper proposes \textbf{DecoFuse}, a novel framework for fMRI-to-video decoding that addresses the challenge of reconstructing videos from brain activity by decomposing the video into three key components: semantic, spatial, and motion information.
\textbf{(2) \textit{Novel Designs of Various Encoders and Decoders}}. Our DecoFuse nontrivially improves upon previous works, featuring novel fMRI, semantic, spatial, and motion encoders.
\textbf{(3) \textit{Biologically Plausible Design}}: DecoFuse's modular approach closely aligns with the two-streams hypothesis. Our ablation studies demonstrate a strong correlation between the learned representations and their biological counterparts. 
\textbf{(4) \textit{ Differential Neural Encoding}}: Investigates the alignment of decoded embeddings with the brain's dorsal and ventral streams; and uses PCA and ridge regression to predict fMRI signals from semantic and spatial embeddings. Essentially, it supports the well established neuroscience theories.
%
%
\textbf{(5) \textit{Superior Performance}}: DecoFuse significantly outperforms state-of-the-art methods in decoding semantic, spatial, and motion components.







\section{Related Work}

\noindent \textbf{fMRI-to-vision reconstruction}.
Recent advances in fMRI-based decoding have made significant strides in extracting visual information from brain activity, particularly in decoding images, videos, and 3D objects using techniques like Stable Diffusion (SD)~\cite{mind-vis, huo2025neuropictor, lienhancing, mind-video, gong2024neuroclips, mind-3d}. However, fMRI-to-video decoding remains underexplored, especially in terms of spatial and motion components. Early works~\cite{mind-video,lienhancing,sun2024neurocine} focused primarily on semantic decoding, while more recent approaches~\cite{lu2024animate, foscobrain,gong2024neuroclips,yeung2024neural} have incorporated VAE latent or motion-specific decoders. Nonetheless, evaluations have typically concentrated on semantic or pixel-level metrics, leaving the reliable decoding of spatial and motion information as an ongoing challenge.

\noindent \textbf{Visual pathways in brain}.
Numerous studies in neuroscience have explored how the brain processes visual information. The two-streams hypothesis~\cite{GOODALE199220, milner2006visual} proposes that visual processing is divided into two pathways: the ``what'' pathway (ventral stream) for object recognition, and the ``where''/``how'' pathway (dorsal stream) for tracking object location and movement. These pathways correspond to the three components of video\textemdash semantic (what), spatial (where), and motion (how)\textemdash which are crucial for reconstructing realistic video content. 

\section{Method}


\noindent \textbf{Overview}.
We decompose the task into  semantic, spatial, and motion decoding, respectively. In data preprocessing, raw fMRI frames are aligned with an anatomical brain template~\cite{glasser2013minimal} to create single-channel images. These fMRI frames are then fed into a large-scale fMRI Pretrained Transformer Encoder (fMRI-PTE)~\cite{qian2023fmri}, which is pretrained on the UKB~\cite{UKB} dataset. Next, two independent modules separately decode the semantic and spatial embeddings, producing a single image via Stable Diffusion~\cite{SD}. Finally, using both the fMRI data and the generated image, a motion decoder predicts optical flow, and DragNUWA~\cite{yin2023dragnuwa} animates the static object in the image to generate the video.

Generally, 
combining the two-streams hypothesis (``what'' and ``where'' concepts) with a brain decoding model offers new insights. 
Building on this, our brain-inspired method links deep learning embeddings to the brain's encoding process, helping us analyze brain signals more effectively by separating different variables.




\subsection{Data Pre-processing}

\noindent\textbf{fMRI preprocessing}.
Some decoding methods flatten each frame and intentionally filter subject-specific activated voxels~\cite{Wen2017,mind-video}. In contrast, we align the fMRI data to the fs\_LR\_32k brain surface space using anatomical structures~\cite{glasser2013minimal} and unfold the cortical surface to create a 2D image, ensuring a standardized and unified representation across subjects while preserving spatial relationships between adjacent voxels. Given that visual tasks primarily activate specific brain regions~~\cite{huang2021fmri}, we concentrate on early and higher visual cortical Regions of Interest (ROIs) covering 8,405 vertices, as defined by the HCP-MMP atlas~\cite{Glasser2016AMP} in the fs\_LR\_32k space. Each fMRI frame is then transformed into a one-channel 256$\times$256 image, followed by voxel-wise z-transformation. Additionally, temporally aligned fMRI frames from different runs with the same video stimulus are averaged. Finally, we apply an approximate 6-second temporal shift to the fMRI series considering the inherent time lag between the stimulus input and the peak of the BOLD signal due to the hemodynamic response.

\noindent\textbf{fMRI-stimuli paired data}. We follow the MinD-Video~\cite{mind-video} and use a sliding window approach to split the CC2017 dataset~\cite{Wen2017} into fMRI-video paired samples. Specifically, the fMRI-to-video decoding task is reformulated as generating a $T$-second video from $\alpha T$-seconds of fMRI data. Additionally, inspired by the two-streams hypothesis~\cite{GOODALE199220}, which suggests that ``what'', ``where'', and ``how'' information is primarily encoded by different brain regions, we decompose the video to semantic, spatial and motion components. These are represented by the initial frame (semantic and spatial) and optical flow (motion).

Assuming there are $n$ frames of fMRI $\mathbf F_i\in \mathbb{R}^{n\times H_f\times W_f}$ and $m$ frames of video $\mathbf V_i\in\mathbb{R}^{m\times 3\times H_v\times W_v}$ in the $i$-th window, where $H_f,W_f$ and $H_v,W_v$ denotes the height and width of the unfolded fMRI image and video. Each optical flow $\mathbf O_i^{k}\in \mathbb{R}^{H_v\times W_v\times 2}$ is then generated by MemFlow~\cite{dong2024memflow} using the initial frame $\mathbf V_i^{k}$ and the future frame $\mathbf V_i^{k+\lfloor\frac{m}{2}\rfloor}$, which can be formulated as
\begin{align}
\mathbf O_i^k=\mathrm{MemFlow}(\mathbf V_i^k,\mathbf V_i^{k+\lfloor \frac{m}{2}\rfloor})
\end{align}
where $1\leq k\leq \lfloor\frac{m}{2}\rfloor$.

\subsection{DecoFuse pipeline}



Visual input is essential for the brain, and many studies have explored how it processes this information.
The well-known two-streams hypothesis suggests that the brain processes visual information through two distinct pathways: the ``what'' pathway (ventral stream) for recognizing objects and the ``where/how'' pathway (dorsal stream) for tracking their location and movement~\cite{GOODALE199220, milner2006visual}. Motivated by this, we propose a brain-inspired fMRI-to-video framework, \textbf{DecoFuse}, which decomposes a video into three components: semantic (``what''), spatial (``where''), and motion (``how''), separately decodes each component, and finally fuses them to generate the video.

\noindent \textbf{fMRI encoder}. To reduce information loss when encoding high-dimensional fMRI signals into a compact feature space, we apply fMRI-PTE~\cite{qian2023fmri}, a ViT-based autoencoder pretrained on a large-scale fMRI dataset~\cite{UKB}, as our encoder. Unlike those ViT-based encoders that flatten and patchify voxels without preserving spatial information, this approach retains local structure~\cite{mind-vis,mind-video}. Each 2D fMRI frame $\mathbf F_i^t\in\mathbb{R}^{H_f\times W_f}$ is divided into $p$ square patches, where each patch represents a token that captures the spatial relationships between neighboring voxels. These patchified fMRI images are then transformed into token embeddings $\mathbf F^{t}_{emb,i}\in\mathbb R^{(p+1)\times D_f}$ through a series of spatial attention blocks, with $D_f$ representing the embedding dimension. The model achieves high-precision reconstruction using only the [CLS] token, yielding an encoder that effectively retains the main information.

\noindent \textbf{Stage 1: semantic and spatial decoding}. 
In this stage, we decode semantic and spatial information from fMRI data to reconstruct static keyframes. Recent advances in image editing~\cite{controlnet} demonstrate that high-level semantic latent codes can guide the semantic content of generated images, while updating feature maps allows precise control over spatial composition. Building on this insight, as illustrated in \cref{fig:method}, we employ an fMRI-to-image pipeline that integrates semantic guidance and spatial control to enhance Stable Diffusion (SD)~\cite{SD}. Based on the high-level and low-level framework from NeuroPictor~\cite{huo2025neuropictor}, our approach further deepens the encoding process and augment the semantic encoder to improve decoding performance.



For high-level semantic decoding, we use a semantic encoder \(\mathcal{E}_{sem}\) to transform fMRI features \(\mathbf{F}_{emb}\) into semantic embeddings \(\mathbf{E}_{sem} = \mathcal{E}_{sem}(\mathbf{F}_{emb})\), replacing the typical text embeddings \(\mathbf{E}_{txt}\) in Stable Diffusion, where \(\mathbf{E}_{sem}, \mathbf{E}_{txt} \in \mathbb{R}^{L_T \times D_T}\). Unlike NeuroPictor, which uses convolutional layers and MLPs in its encoder, we use transformer layers to capture semantic information related to the visual stimulus. This helps guide the diffusion model, ensuring the generated image accurately reflects the perceived objects and scene context.


For spatial decoding, we use a spatial encoder \(\mathcal{E}_{spa}\) to directly adjust the feature maps in the U-Net architecture of the diffusion model. The spatial embeddings are derived as \(\mathbf{E}_{spa} = \mathcal{E}_{spa}(\mathbf{F}_{emb})\), where \(\mathbf{E}_{spa} = \{\mathbf{E}_{spa,(i)} \mid i = 1, \ldots, 13\}\), with \(\mathbf{E}_{spa,(i)}\) representing the feature map from the \(i\)-th encoder block. The spatial encoder applies channel-wise convolutions, MLPs, and transformer layers to refine U-Net feature maps at various levels. The resulting spatial embeddings are processed through zero convolution layers and combined with the intermediate outputs of the SD model using a residual connection:
\begin{align}
\tilde{\mathbf{E}}_{spa} = \mathbf{E}_{SD} + \alpha \mathcal{Z}(\mathbf{E}_{spa})
\end{align}
where \(\mathcal{Z}\) is the zero convolution layer, \(\mathbf{E}_{SD}\) represents the latent codes of the SD U-Net, and \(\alpha\) is a hyperparameter balancing high-level semantic guidance and fine-grained spatial details. This method effectively controls detailed spatial features, such as object positioning and structural layout.

By combining the semantic guidance \(\mathbf{E}_{sem}\) and spatial guidance \(\tilde{\mathbf{E}}_{spa}\) derived from fMRI, we can finely control the generated outputs, achieving both semantic and spatial reconstruction of static images.

\noindent \textbf{Stage 2: motion decoding}. 
Previous work~\cite{yeung2024neural} has demonstrated that motion information, such as optical flow, can be decoded from fMRI. Therefore, we propose a motion decoder that predicts optical flow of a video based on fMRI and its first frame. In other word, motion decoder functions by ``asking'' the frozen brain (fMRI) how objects in the first frame are moving in the viewed video. Moreover, we suggest that in a short video (e.g., a 2-second clip), only coarse movement can be reliably encoded in fMRI due to its low temporal and spatial resolution. As a result, our motion decoder $\mathcal D_M$ predicts only a single frame of low-resolution optical flow for each sample.
\begin{align}
\hat{\mathbf O}_i^k=\mathcal D_M(\mathbf V_i^k,\mathbf F_{i})
\end{align}

To accurately decode motion information, we follow prior image-to-motion work~\cite{walker2015dense}, which showed that optical flow classification outperforms direct prediction. First, we flatten the vectors from all optical flow $\mathbf O_i$ in training set and apply K-means clustering to obtain a codebook $\mathbf B\in\mathbb R^{N_{vec}\times 2}$, where $N_{vec}$ is the number of clusters. Each vector in the optical flow $\mathbf O_i$ is then quantized by its nearest vector in the codebook. The quantized optical flow $\tilde{\mathbf O}$ is defined as:
\begin{align}
\tilde{\mathbf{O}}_{i,h,w}^k=\mathbf B_{c^\star},\ \  c^\star=\arg\min_{c} \parallel\mathbf O_{i,h,w}-\mathbf B_c\parallel_2^2
\end{align}

More specifically, $\mathbf V_i^k$ and $\mathbf F_i$ are sent to their corresponding pretrained encoders, DINOv2~\cite{oquab2023dinov2} and fMRI-PTE~\cite{qian2023fmri} to generate token-level embeddings. As shown in \cref{fig:method} (Stage 2), after separate CNN processing, the two embeddings are concatenated and passed through a CNN and softmax layer to predict probability distribution $\mathbf P_i^k\in\mathbb R^{H_o\times W_o\times N_{vec}}$ of vectors in codebook. The final prediction of optical flow is then given by $\hat{\mathbf O}_i^k=\mathbf{P}_i^k\mathbf B$. 

\noindent \textbf{Stage 3: video generation}.
Based on the pre-generated image and optical flow, we reconstruct the video using DragNUWA~\cite{yin2023dragnuwa}, a pretrained video diffusion model conditioned on motion. First, to ensure more stable video generation, we mask the optical flow using foreground detection from DINOv2~\cite{oquab2023dinov2}. Additionally, to generate an $N_f$-frame video, we extend the single-frame optical flow by linearly dividing the vector to $N_f-1$ sub-vectors. 

\subsection{Differential neural encoding}

Since DecoFuse is inspired by the two-streams hypothesis~\cite{GOODALE199220}, we conduct neural encoding to examine whether and how the decoded embeddings differentially align with the two streams identified in biological studies. To prevent overfitting, we first apply Principal Component Analysis (PCA) to reduce the dimension of the semantic embedding $\mathbf E_{sem}$ and spatial embedding $\mathbf E_{spa}$, resulting in $\mathbf E^{PCA}_{sem},\mathbf E^{PCA}_{spa}\in\mathbb R^{T\times D}$, where $T$ is the number of time points in the fMRI volumes, and $D$ is the reduced dimension. We then use ridge regression to predict the Gaussian-smoothed and flattened fMRI data $\mathbf F\in\mathbb R^{T\times N_v}$ based on semantic or spatial embeddings, where $N_v$ represents the number of voxels.
\begin{align}
\hat{\mathbf F}_{X}=\mathrm{RidgeRegressor}(\mathbf E^{PCA}_{X}), X\in\{sem,spa\}
\end{align}
Next, we compute the average temporal correlation $\mathbf r_X$ for the predicted and GT fMRI signals over a window size $T_w$:
\begin{align}
\mathbf r_{X}=\frac{1}{N_w}\sum_{t=1}^{N_w} \mathrm{corr}(\mathbf F_{X,t:t+T_w}, \hat{\mathbf F}_{X,t:t+T_w}),\nonumber\\
 X\in\{sem,spa\}
\end{align}
Following metrics in~\cite{choi2023dual}, we differentiate the relative contributions of the semantic and spatial embeddings in predicting the brain's dorsal and ventral streams for individual voxels:
\begin{align}
\mathbf p_{spa}=\frac{\mathbf r_{spa}^2}{\mathbf r_{spa}^2+\mathbf r_{sem}^2}-0.5
\end{align}
Here, $\mathbf p_{spa}$ ranges from -0.5 to 0.5. A value of $p_{spa}>0$ indicates that the spatial embedding better predicts the voxel, while $p_{spa}<0$ suggests that semantic embedding provides a better prediction.

\subsection{Training Strategy}

We perform training in both Stage 1 and Stage 2.

\noindent \textbf{Stage 1.} We freeze the SD model to retain its strong image synthesis capabilities, while finetuning the semantic, spatial, and fMRI encoders to extract semantic and spatial information from fMRI data. Since the image represents static information, we use a single fMRI frame for image decoding. The pipeline is trained with fMRI-image pairs $(\mathbf F_i^1,\mathbf V_i^k)$, where $1\leq k\leq\lfloor\frac{m}{2}\rfloor$ denotes data augmentation for random initial frame.

Specifically, the input image $\mathbf V_i^k$ is first encoded into a latent representation $z_0$. The diffusion process then progressively adds noise to $z_0$ over $t$ time steps, resulting in a noisy latent $z_t$. During the denoising stage, the frozen U-Net predicts a denoised version of $z_t$, conditioned on the time step $t$, semantic embedding $\mathbf E_{emb}$, and spatial embedding $E_{spa}$. The denoising loss for optimizing the SD latent is defined as follow:
\begin{align}
\mathcal L_{s1}=\mathbb E_{z_0,t,\mathbf F_{emb},\epsilon\sim\mathcal N(0,1)}\Big[\parallel \epsilon-\epsilon_\theta (z_t,t,\mathbf F_{emb},\mathbf E_{emb})\parallel^2_2 \Big]
\end{align}

\noindent  \textbf{Stage 2.} For training the motion decoder, we use fMRI-image-motion paired data $(\mathbf F_i,\mathbf V_i^k, \mathbf O_i^k)$, where $1\leq k\leq \lfloor\frac{m}{2}\rfloor$ denotes data augmentation for random frame. We combine cross-entropy loss $\mathcal L_{entropy}$ and mean squared error (MSE) loss $\mathcal L_{MSE}$ to form the total loss $\mathcal L_{s2}=\mathcal L_{entropy}+\lambda_2\mathcal L_{MSE}$ for training the motion decoder $\mathcal D_M$.
\begin{align}
\mathcal L_{entropy}&=\mathrm{CrossEntropy}(\mathbf P_i^k,\mathbf c_i^k)\\
\mathcal L_{MSE}&=\parallel \mathbf O_i^k-\hat{\mathbf O}_i^k\parallel_2^2
\end{align}
where $\mathbf c_i^k$ is codebook label for optical flow $\mathbf O_i^k$.

\section{Experiments}


\noindent\textbf{Pre-training Dataset}.
The UK Biobank (UKB) \cite{UKB} is a large-scale biomedical resource that gathers extensive genetic and health-related data from roughly 500,000 individuals across the UK. A subset of this repository is utilized, specifically the resting-state fMRI data from approximately 39,630 participants. Each participant provides a single session consisting of 490 time-point volumes.


\noindent\textbf{Paired fMRI-video Dataset}. 
Experiments used the CC2017 dataset \cite{Wen2017}, which pairs fMRI data with video stimuli. It includes data from three participants, with fMRI frames captured using a 3T MRI scanner at a 2-second repetition time (TR). The dataset covers about 3 hours of video and provides around 5,500 fMRI-stimulus pairs per subject.


\noindent \textbf{Vision metrics}.
\textbf{1) Semantic-level}. Following Mind-Video~\cite{mind-video}, we use both image-based and video-based classification metrics to assess semantic-level performance. For image classification, we rely on ImageNet classifier. For the video-based metrics, we apply a similar classification framework, utilizing VideoMAE \cite{videomae}. In both cases, the N-way top-K accuracy metric is employed, where for video the top-3 predicted classes are compared against the ground truth (GT) class. Specifically, N candidates include the ground truth class along with N-1 randomly selected classes from the classifier's full class set. This approach is consistent with the methodology used in MinD-Video. \textbf{2) Spatial-level}. We evaluate spatial performance by calculating the ratio of foreground-background matching between the ground truth and decoded images. Foreground detection is performed using DINOv2~\cite{oquab2023dinov2}. Let the matrix $\mathbf M\in \{0,1\}^{H\times W}$ represent the foreground mask, where $\mathbf M_{i,j}=1$ indicates pixel $(i,j)$ is detected as foreground, and $\mathbf M_{i,j}=0$ indicates background. The matching ratio $r_m$ is then calculated as follows:
\begin{align}
r_m=1-\frac{\parallel \mathbf M_{GT}-\mathbf M_{pred}\parallel_0}{H\times W}
\end{align}
where $\mathbf M_{GT}, \mathbf M_{pred}$ represent foreground mask metrics of GT and predicted images, respectively. The value of $r_m$ ranges from 0 to 1, with a value closer to 1 indicating better matching of the foreground and background between ground truth and predicted images, and a value closer to 0 indicating worse matched.
\textbf{3) Pixel-level}. We use the structural similarity index measure (SSIM)~\cite{SSIM} to assess pixel-level decoding performance. For video evaluation, SSIM is computed for each frame of both the ground truth and reconstructed videos, with results averaged across frames. 

\noindent  \textbf{Motion metrics}.
We evaluate motion decoding performance using cosine similarity between the ground truth and decoded optical flow vectors. To handle scene changes that may produce invalid optical flow, we apply scene-change detection to remove such samples. Additionally, we mask all predicted optical flow using foreground detection and also mask ground truth values close to the zero vector to reduce noise. Specifically, the shortest cluster in the quantized codebook is set to the zero vector.


\noindent  \textbf{Implementation Details}.  For CC2017~\cite{Wen2017}, we used an fMRI window of $\alpha$T = 2s to generate videos lasting T = 2s and videos were downsampled to 8 FPS. All training and inference processes were conducted on a single NVIDIA A100 GPU. Please refer to the Supplementary for detailed hyperparameter configurations and additional training information. Codes\&Models will be released.

\subsection{Verifying the  `What' and `Where' Factor}

\input{tables/fMRI-to-image-result}

\begin{figure}[t]
\centering
\includegraphics[width=0.95\linewidth]{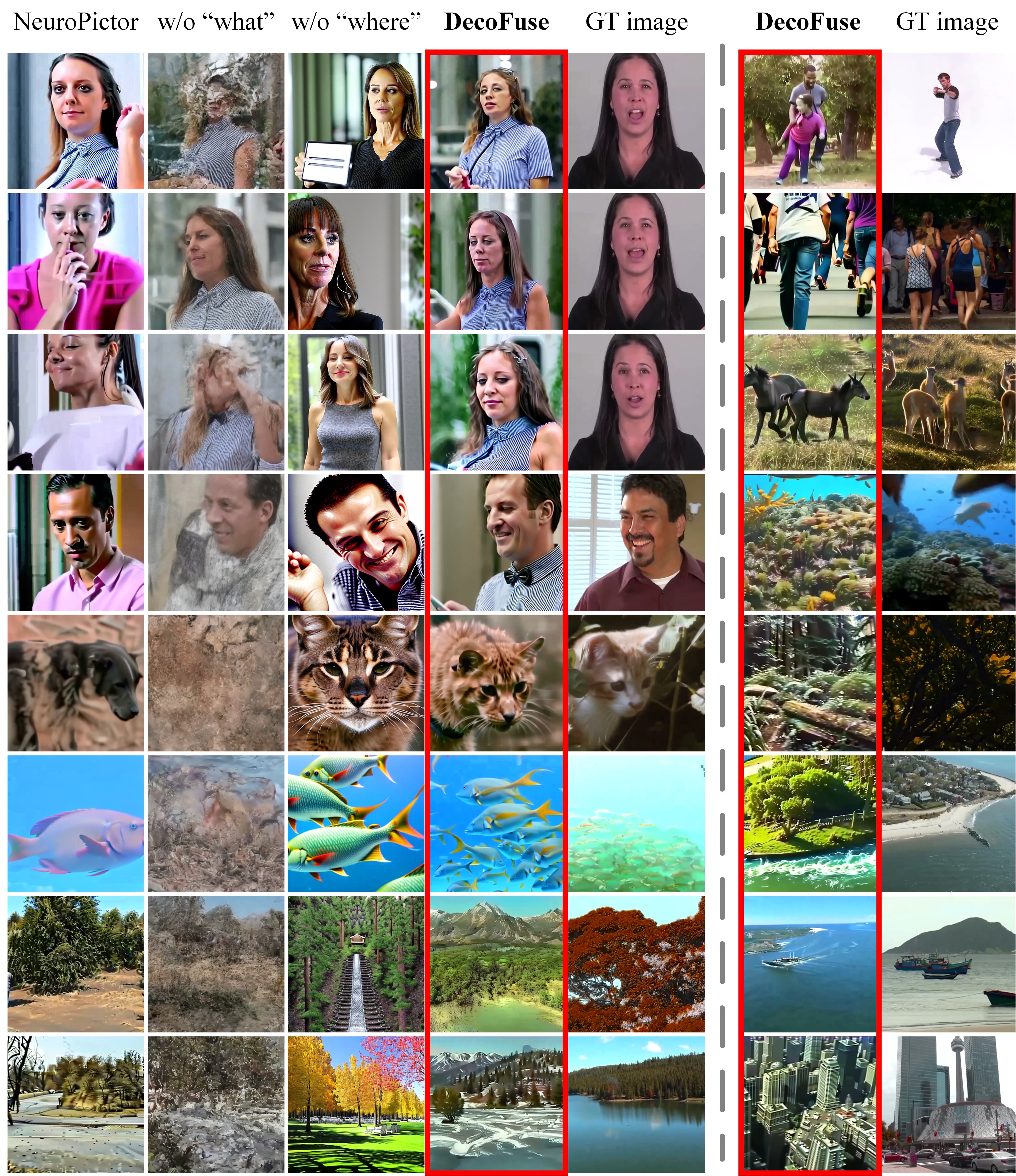}
\vspace{-0.1in}
\caption{\textbf{Results of fMRI-to-image reconstruction.} Our model successfully generates images that align well with the ground truth in both semantic and spatial aspects. By comparing the results with and without semantic(``what'')/spatial(``where'') embeddings, we demonstrate that semantic and spatial embeddings significantly enhance the model’s ability to accurately reconstruct and localize objects within the image.\label{fig:fMRI-to-image-result} }

\vspace{-0.2in}
\end{figure}

To isolate the ``What'' and ``Where'' components in decoded images from fMRI signals, we compare our method, DecoFuse, with other established fMRI-to-video decoding approaches, including MinD-Video~\cite{mind-video}, fMRI-PTE-video~\cite{lienhancing}, and NeuroPictor~\cite{huo2025neuropictor}.

Our findings, as in \cref{fig:fMRI-to-image-result} and \cref{tab:fMRI-to-image-result}, show results across three subjects with both semantic and spatial metrics. (1) DecoFuse consistently outperforms the other methods in these metrics, capturing detailed semantic content and accurately decoding spatial locations. This shows DecoFuse's ability to better align  ``What'' (semantic content) and ``Where'' (spatial arrangement) from brain activity, setting a new benchmark in fMRI-to-video decoding. (2) At the semantic level, DecoFuse achieves significantly higher accuracy than the other methods. For example, in subject 1, DecoFuse achieves a 50-way accuracy of 0.208, compared to 0.172 for MinD-Video, 0.169 for fMRI-PTE-video, and 0.195 for NeuroPictor. This trend holds across subjects, with DecoFuse leading in both 2-way and 50-way accuracy, demonstrating its effectiveness in capturing semantic content from fMRI data. (3) At the spatial level, DecoFuse excels in preserving spatial locations. For instance, in subject 1, DecoFuse achieves a matching ratio of 0.706, outperforming MinD-Video (0.660), fMRI-PTE-video (0.652), and NeuroPictor (0.687), indicating better object localization.


To evaluate the impact of semantic and spatial features, we ablate these embeddings in DecoFuse respectively. DecoFuse\text{\scriptsize{(w/o where)}}, which excludes spatial features, shows a clear drop in spatial metrics, confirming their importance. DecoFuse\text{\scriptsize{(w/o what)}}, which removes semantic conditioning, experiences a significant decline in semantic accuracy but retains a high spatial score of 0.704. Additionally, to reduce randomness, DecoFuse generates 20 frames and selects the one with the least deviation (see Supplementary for details), while DecoFuse\text{\scriptsize{(1 frame)}} generates only a single frame. The results show that filtering one frame from multiple frames improves performance by reducing generation variance.
Overall, DecoFuse excels in both semantic and spatial decoding, capturing fine fMRI details and generating high-quality visual reconstructions, surpassing previous methods.


\subsection{Verifying the `How' factor}

\begin{figure*}
\centering
\includegraphics[width=0.85\linewidth]{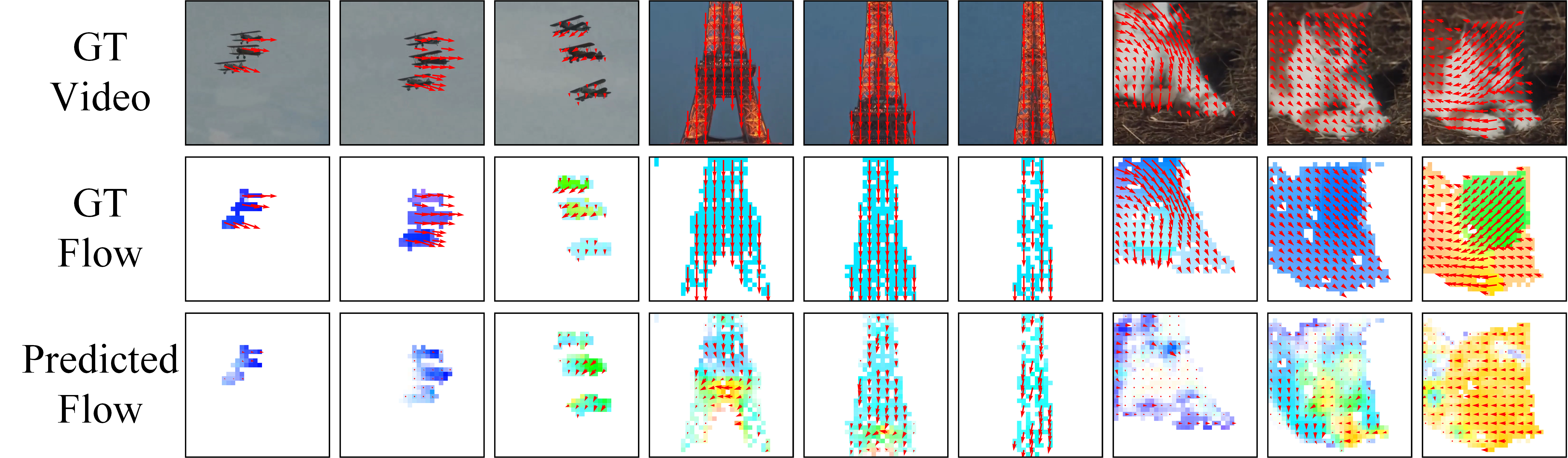}
\vspace{-0.1in}
\caption{\textbf{Results of fMRI-to-motion decoding.} Our model effectively predicts optical flow based on fMRI and image data, demonstrating accurate motion decoding performance.\label{fig:fMRI-to-motion-result} }

\vspace{-0.15in}
\end{figure*}

\input{tables/fMRI-to-motion-result.tex}

\noindent \textbf{`Disclaimer'}. Since there is no direct way to make a fair comparison for the ``How'' factor, we adapt optical flow metrics for evaluation. However, optical flow is highly sensitive to various factors—occlusions, rapid motion and motion blur, changes in illumination, and even noise or artifacts—all of which commonly appear in generated images of all methods. As a result, it is challenging to quantify the exact impact these sensitivities might have on our comparisons. Nonetheless, optical flow still provides a useful baseline metric, offering a general gauge for assessing the effectiveness of each method.

To assess motion decoding performance, we measure cosine similarity between predicted and ground truth optical flow vectors across varying foreground coverage levels. In \cref{tab:fMRI-to-motion-result}, each percentile (e.g., 20\%, 30\%, etc.) represents the proportion of the scene occupied by the foreground, offering insights into how well each model decodes motion with emphasis on larger, more prominent objects. This approach reflects the human tendency to focus on movement associated with larger scene elements.

The motion decoding results in \cref{fig:fMRI-to-motion-result} and \cref{tab:fMRI-to-motion-result} demonstrate DecoFuse's capabilities relative to the fMRI-to-motion (F2M) method~\cite{yeung2024neural}, using cosine similarity across these foreground thresholds. Although exact comparisons are limited by the F2M algorithm's incomplete details, DecoFuse presents a notable edge. For instance, our method’s computation of optical flow at one-second intervals introduces added complexity, yet DecoFuse still demonstrates strong performance.  In particular, DecoFuse excels in capturing motion within larger foreground regions, outperforming F2M. This pattern supports our hypothesis that DecoFuse aligns closely with human perceptual biases, effectively prioritizing motion decoding for visually dominant areas. These results affirm DecoFuse’s robust motion decoding ability, especially in challenging conditions that require precision with significant scene elements.

We also tested optical flow prediction after ablating fMRI input, which is equivalent to optical flow prediction based only on images. The results show that predictions based solely on images perform much worse compared to predictions made with both fMRI and images. This suggests that the model successfully learns motion information from the fMRI data.

\subsection{More Ablation Study}

\begin{figure}
\centering
\includegraphics[width=0.95\linewidth]{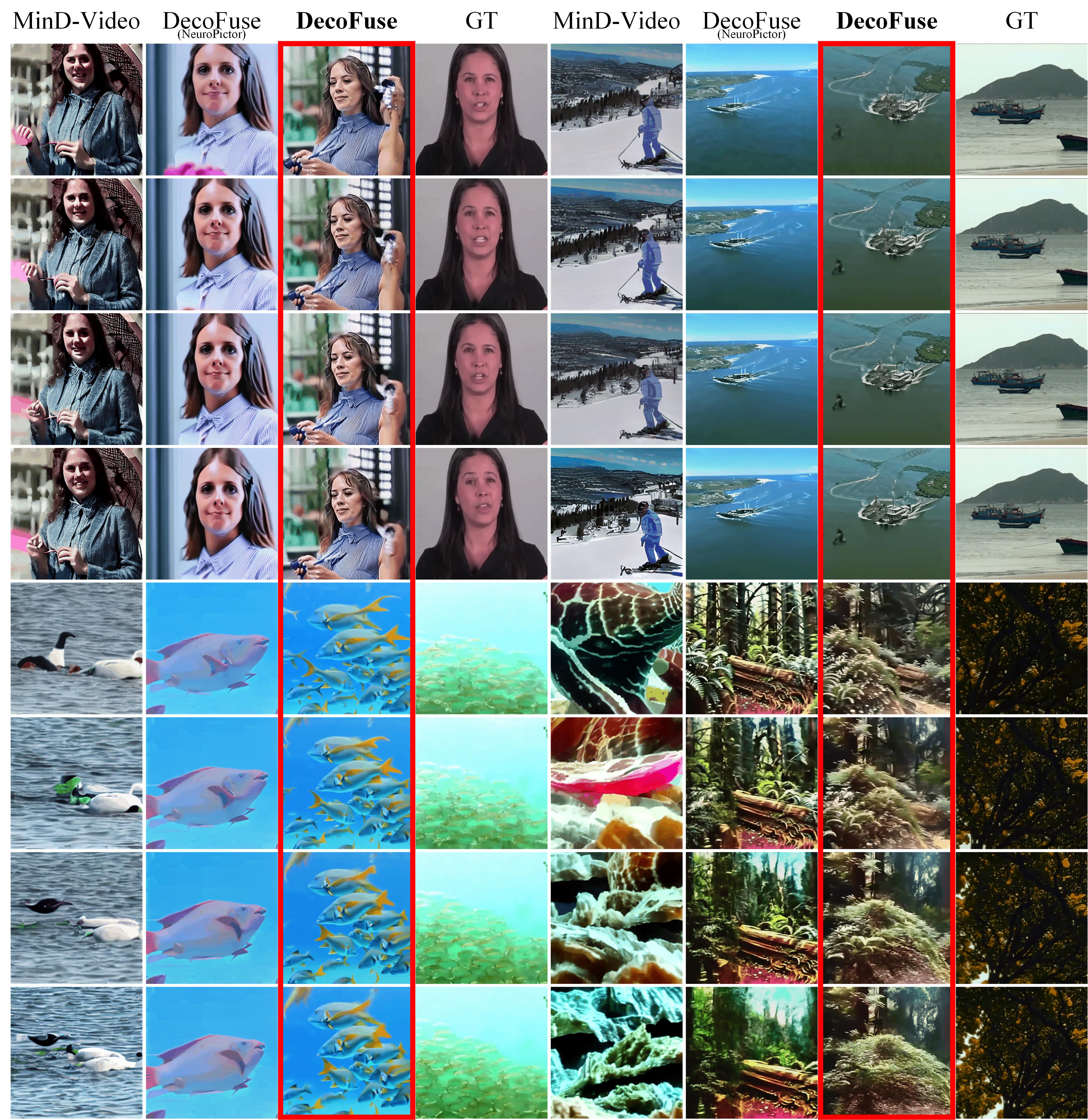}
\vspace{-0.1in}
\caption{\textbf{Our fMRI-to-video decoding.} Our model shows accurate decoding performance at both the semantic and pixel levels. \label{fig:fMRI-to-video-result} }

\vspace{-0.15in}
\end{figure}

\noindent \textbf{Other impacting factors in decoding videos}. 
We further evaluate the direct decoding of videos from fMRI by semantic-level accuracy and structural similarity (SSIM), following the metrics used in~\cite{mind-video}. For each subject, we report both 2-way and 50-way semantic accuracy. As shown in \cref{tab:fMRI-to-video-result}, DecoFuse demonstrates best performance on most cases, highlighting the improved accuracy of our decoded videos. These results affirm DecoFuse's effectiveness in preserving both semantic and structural details from fMRI data. We also provide visualizations of the decoded frames in \cref{fig:fMRI-to-video-result}, highlighting the clarity and fidelity of our approach. Additionally, we assess video decoding (DecoFuse\text{\scriptsize{(NeuroPictor)}}) based on images generated by NeuroPictor~\cite{huo2025neuropictor}, showing a significant decrease in semantic metrics, which further proves the improvement of our fMRI-to-image decoding pipeline.

\input{tables/fMRI-to-video-result.tex}

\begin{figure}
\centering
\includegraphics[width=1\linewidth]{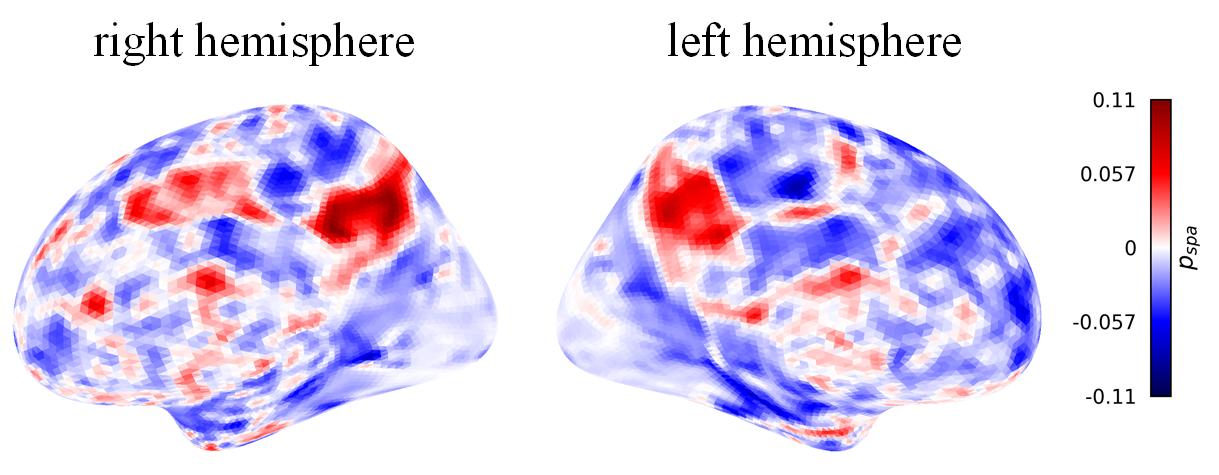}
\vspace{-0.2in}
\caption{\textbf{Results of differential neural encoding.} The differential encoding distribution for ``what'' and ``where'' is represented by $p_{spa}$ and visualized on the medial view of the brain surface. Red indicates regions that encode ``where'' information, while blue indicates regions that encode ``what'' information. These results align with the two-streams hypothesis~\cite{GOODALE199220}.\label{fig:neural-encoding} }
\vspace{-0.15in}

\end{figure}

\noindent \textbf{Differential Neural Encoding}.
We explore how the brain encodes semantic and spatial information through distinct pathways using differential neural encoding,as visualized in \cref{fig:neural-encoding}. It highlights the contributions of semantic and spatial features in predicting fMRI responses.  So our findings support the two-streams hypothesis~\cite{GOODALE199220}. In the primary visual cortex, when $p_{spa}$ approaches 0, both types of information are encoded equally. As processing progresses through the dorsal and ventral pathways, a bias emerges, favoring spatial or semantic cues, respectively. In higher-order regions, such as the frontal lobe, this distinction diminishes, supporting the idea that our approach decodes brain activity in a manner consistent with biological encoding processes.

\section{Conclusion}
This paper introduces DecoFuse, a novel fMRI-to-video decoding framework that separates video into semantic, spatial, and motion components. By independently decoding these aspects, DecoFuse provides a more accurate reconstruction of visual experiences, addressing the brain's ``what'', ``where'', and ``how'' pathways. Unlike existing methods focused on semantic information, DecoFuse incorporates spatial and motion components for more realistic video reconstruction.

{
\small
\bibliography{arxiv}
}

\end{document}

%% file: tables/fMRI-to-image-result.tex
\setlength{\tabcolsep}{2.5pt} 
\begin{table}[]
	\centering
	\begin{tabular}{ccccc}
		\hline
		\multirow{2}{*}{}     & \multirow{2}{*}{Methods}       & \multicolumn{2}{c}{Semantic-level} & Spatial-level  \\ \cline{3-5} 
		&                                       & 2-way          & 50-way         & $r_m$          \\ \hline
		\multirow{7}{*}{sub1} & MinD-Video~\cite{mind-video}   & 0.792            & 0.172           & 0.660          \\
		& fMRI-PTE-video~\cite{lienhancing}     & 0.793          & 0.169          & 0.652          \\
		& NeuroPictor~\cite{huo2025neuropictor} & 0.808          & 0.195          & 0.687          \\ \cline{2-5} 
		& DecoFuse\scriptsize{(w/o what)}       & 0.774          & 0.130          & \textbf{0.704} \\
		& DecoFuse\scriptsize{(w/o where)}      & 0.792          & 0.171          & 0.668          \\
		& DecoFuse\scriptsize{(1 frame)} & \textbf{0.816}   & \textbf{0.201}  & \textbf{0.690} \\
		& DecoFuse                              & \textbf{0.824} & \textbf{0.208} & \textbf{0.706} \\ \hline
		\multirow{4}{*}{sub2} & MinD-Video                            & 0.784          & 0.158          & 0.669          \\
		& fMRI-PTE-video                        & 0.780          & 0.159          & 0.648          \\
		& NeuroPictor                           & 0.785          & 0.169          & 0.679          \\
		& DecoFuse                              & \textbf{0.802} & \textbf{0.190} & \textbf{0.692} \\ \hline
		\multirow{4}{*}{sub3} & MinD-Video                            & 0.812          & 0.193          & 0.662          \\
		& fMRI-PTE-video                        & 0.799          & 0.173          & 0.637          \\
		& NeuroPictor                           & 0.803          & 0.194          & 0.671          \\
		& DecoFuse                              & \textbf{0.816} & \textbf{0.215} & \textbf{0.689} \\ \hline
	\end{tabular}
	\vspace{-0.1in}
	\caption{\textbf{Results of fMRI-to-image decoding.} Evaluations of semantic and spatial metrics are presented for all three subjects, with bolded results indicating performance surpassing all baselines. \label{tab:fMRI-to-image-result}}
	
	\vspace{-0.15in}
	
\end{table}

%% file: tables/fMRI-to-motion-result.tex

\setlength{\tabcolsep}{2pt} 
\begin{table}[]
	\centering
	\begin{tabular}{ccccccc}
		\hline
		\multirow{2}{*}{}     & \multirow{2}{*}{Methods}        & \multicolumn{5}{c}{cosine similarity}                                     \\ \cline{3-7} 
		&                                 & 20\%  & 30\%           & 40\%           & 50\%           & 60\%           \\ \hline
		\multirow{3}{*}{sub1} & F2M~\cite{yeung2024neural}      & \multicolumn{5}{c}{0.174}                                                 \\ \cline{2-7} 
		& DecoFuse\scriptsize{(w/o fMRI)} & 0.051 & 0.049          & 0.026          & 0.016          & -0.042         \\
		& DecoFuse                        & 0.139 & 0.147          & 0.150          & \textbf{0.212} & \textbf{0.179} \\ \hline
		\multirow{2}{*}{sub2} & F2M                             & \multicolumn{5}{c}{0.085}                                                 \\ \cline{3-7} 
		& DecoFuse                        & 0.045 & 0.052          & 0.055          & -0.028         & -0.137         \\ \hline
		\multirow{2}{*}{sub3} & F2M                             & \multicolumn{5}{c}{0.110}                                                 \\ \cline{3-7} 
		& DecoFuse                        & 0.106 & \textbf{0.129} & \textbf{0.153} & \textbf{0.144} & 0.050          \\ \hline
	\end{tabular}
	\vspace{-0.1in}
	\caption{\textbf{Results of fMRI-to-motion decoding.} The details of how F2M~\cite{yeung2024neural} computes cosine similarity are not provided. Therefore, we evaluate our method on optical flow where the foreground occupies more than various ratios. \label{tab:fMRI-to-motion-result} }
	
	\vspace{-0.2in}
\end{table}


%% file: tables/fMRI-to-video-result.tex
\setlength{\tabcolsep}{3pt} 
\begin{table}[]
	\centering
	\begin{tabular}{ccccc}
		\hline
		& \multirow{2}{*}{Methods}           & \multicolumn{2}{c}{Semantic-level} & Pixel-level    \\ \cline{3-5} 
		&                                    & 2-way            & 50-way          & SSIM           \\ \hline
		\multirow{5}{*}{sub1} & LEA~\cite{qian2023joint}           & 0.825            & 0.149           & 0.137          \\
		& MinD-Video~\cite{mind-video}       & 0.853            & 0.202           & 0.171          \\
		& fMRI-PTE-video~\cite{lienhancing}  & 0.851            & 0.214           & 0.193          \\ \cline{2-5} 
		& DecoFuse\scriptsize{(NeuroPictor)} & 0.839            & 0.204           & \textbf{0.370} \\
		& DecoFuse                           & \textbf{0.855}   & \textbf{0.219}  & \textbf{0.339} \\ \hline
		\multirow{4}{*}{sub2} & LEA                                & 0.826            & 0.148           & 0.145          \\
		& MinD-Video                         & 0.841            & 0.173           & 0.171          \\
		& fMRI-PTE-video                     & 0.834            & 0.192           & 0.182          \\
		& DecoFuse                           & \textbf{0.846}   & \textbf{0.193}  & \textbf{0.306} \\ \hline
		\multirow{4}{*}{sub3} & LEA                                & 0.834            & 0.160           & 0.137          \\
		& MinD-Video                         & 0.846            & 0.216           & 0.187          \\
		& fMRI-PTE-video                     & 0.851            & \textbf{0.225}  & 0.176          \\
		& DecoFuse                           & \textbf{0.856}   & 0.218           & \textbf{0.314} \\ \hline
	\end{tabular}
	\vspace{-0.1in}
	\caption{\textbf{Results of fMRI-to-video decoding.} Our method outperforms the baselines in most cases, with bolded results highlighting superior performance over all baselines. 	\label{tab:fMRI-to-video-result} }
	
	\vspace{-0.15in}
\end{table}